\documentclass[JIP,submit]{style/ipsj}

\makeatletter
\def\@uketsuke{}
\def\@euketsuke{}
\def\SHUBETUname{}
\makeatother

\usepackage{adjustbox}
\usepackage{booktabs}
\usepackage{multirow}
\usepackage{graphicx}
\usepackage{subcaption}
\usepackage{url}
\usepackage{microtype}

\title{Evaluating Game Difficulty in Tetris Block Puzzle}

\affiliate{NYCU}{National Yang Ming Chiao Tung University, Hsinchu, Taiwan}
\affiliate{AS}{Academia Sinica, Taipei, Taiwan}

\author{Chun-Jui Wang, Jian-Ting Guo, }{NYCU}
\author{Hung Guei, Chung-Chin Shih, }{AS}
\author{Ti-Rong Wu, }{AS}[tirongwu@iis.sinica.edu.tw]
\author{I-Chen Wu}{NYCU, AS}

\begin{document}

\begin{abstract} 

\textit{Tetris Block Puzzle} is a single-player stochastic puzzle in which a player places blocks on an $8\times8$ grid to complete lines; its popular variants have amassed tens of millions of downloads. 
Despite this reach, there is little principled assessment of which rule sets are more difficult. 
Inspired by prior work that uses AlphaZero as a strong evaluator for chess variants, we study difficulty in this domain using \textit{Stochastic Gumbel AlphaZero (SGAZ)}, a budget-aware planning agent for stochastic environments.
We evaluate rule changes—including holding block $h$, preview holding block $p$, and additional Tetris block variants —using metrics such as training reward and convergence iterations. 
Empirically, increasing $h$ and $p$ reduces difficulty (higher reward and faster convergence), while adding more Tetris block variants increases difficulty, with the T-pentomino producing the largest slowdown.
Through analysis, SGAZ delivers strong play under small simulation budgets, enabling efficient, reproducible comparisons across rule sets and providing a reference for future design in stochastic puzzle games.

\end{abstract}

\begin{ekeyword} Puzzle game, Game rule variants, Stochastic gumbel alpha zero, Game difficulty
\end{ekeyword}
\maketitle






\thispagestyle{empty}
\section{Introduction}



Reinforcement learning (RL) has achieved significant success across a wide range of games.
Notably, AlphaGo~\cite{silver_mastering_2016} employed deep neural networks trained on human game records to achieve superhuman performance in Go. 
Its successor, AlphaGo Zero~\cite{silver_mastering_2017}, removed the need for human data by learning entirely through self-play, surpassing the original AlphaGo. 
Finally, AlphaZero~\cite{silver_general_2018} generalized this approach beyond a single game, achieving state-of-the-art (SOTA) results in many board games, such as Go, chess, and shogi.

After the success of AlphaZero, several extensions have been developed in different directions.
For example, Stochastic AlphaZero~\cite{antonoglou_planning_2021} was introduced to handle stochastic environments where randomness affects state transitions and outcomes.
Meanwhile, because the standard AlphaZero framework provides no guarantee of policy improvement, Gumbel AlphaZero~\cite{danihelka_policy_2022} was proposed to address this limitation.
It ensures policy improvement, allowing effective training even under a very small number of simulation budgets.

Furthermore, as AlphaZero-based algorithms have achieved superhuman performance, they have also been used to explore novel rule sets~\cite{tomašev_assessing_2020} and to control game difficulty~\cite{fujita_alphadda_2022,demediuk_monte_2017}. 
Following this approach, we adopt a similar approach in the \textit{Tetris Block Puzzle}~\cite{website_tetrisblockpuzzle}, a single-player stochastic puzzle game, with Stochastic Gumbel AlphaZero to investigate new gameplay variants and analyze their difficulty.

We investigate the difficulty under different rule variants: the number of holding blocks ($h$), the number of preview holding blocks ($p$), and the inclusion of additional Tetris block types.
We evaluate these variants using two quantitative metrics, including training reward and convergence iterations, to measure how each rule change affects gameplay difficulty.
We observe that larger $h$ and $p$ consistently lower difficulty, yielding higher rewards and faster convergence speed for the strong AI agent. 
Conversely, adding new Tetris block types increases difficulty, most notably the T-pentomino, which produces the largest slowdown in convergence, thereby offering rule configurations that players can select to match their desired challenge.

\section{Background}



\subsection{AlphaZero}



AlphaZero~\cite{silver_general_2018} is a zero-knowledge learning method that achieves superhuman performance in multiple board games without human knowledge.
Training alternates between (i) self-play and (ii) optimization. 
In self-play, each move is selected by running many simulations of Monte Carlo Tree Search (MCTS)~\cite{coulom_efficient_2007} guided by a neural network that outputs a policy and a value. 

Each MCTS simulation performs three phases: selection, expansion, and backpropagation. 
In the selection phase, the search traverses the tree from the root toward a leaf node using a PUCT formula~\cite{schrittwieser_mastering_2020}. 
In the expansion phase, the reached leaf is expanded to the tree, and all legal child nodes are expanded.
The neural network then evaluates the state to provide action priors for those children and a value estimate for the leaf. 
During backpropagation, the value is propagated from the leaf back to the root along the selected path.
After the simulation budget is reached, an action is sampled from the root's action distribution, where the probability of each action is proportional to its number of visits during search.
The finished self-play games are stored in a replay buffer for optimization. 
In the optimization phase, a self-play game is sampled randomly from the replay buffer to update the network.

\subsection{Gumbel AlphaZero}
Although AlphaZero has achieved superhuman performance, it does not guarantee policy improvement when the simulation budget is small. 
The Gumbel Zero framework~\cite{danihelka_policy_2022} was introduced to address this limitation.
It modifies the root selection to improve sample efficiency and guarantee policy improvement.
At the root, it first performs without replacement candidate selection using the \textit{Gumbel-Top-$k$ trick}~\cite{kool_stochastic_2019} on the policy logits, then allocates simulations via sequential halving~\cite{karnin_almost_2013} to progressively prune weaker candidates until a single action remains; that surviving action is taken in the environment deterministically rather than sampling from visit counts. 
For training targets, since only a limited number of root children are explored, visit-count distributions are not used.
Instead, Q-based targets are constructed by using backed-up Q estimates for visited actions and the value network to estimate the values of unvisited ones. 
These modifications preserve the overall AlphaZero training loop while improving efficiency under small simulation budgets and yielding stronger early training performance in practice. 

\subsection{Stochastic AlphaZero}
To extend AlphaZero to stochastic environments, \textit{Stochastic AlphaZero}~\cite{antonoglou_planning_2021} introduces afterstates~\cite{sutton_reinforcement_2018} and incorporates a learned model into the planning process.
The key idea is to separate the effect of the agent's action from the randomness of the environment.
From a given state, applying an action deterministically yields a unique afterstate. 
With the environment's randomness, this afterstate then produces one of several possible successor states according to a probability distribution.

The learned model contains two additional components. 
The afterstate dynamic function $\phi$ predicts the next afterstate $as^k$ given the current state $s^{k-1}$ and chosen action $a^k$. 
The afterstate prediction function $\psi$ estimates, for a given afterstate $as^k$, the set of possible chance outcomes along with their associated probabilities $\sigma^k$ and the afterstate value $Q^k$. 
Then, the dynamic function $g$ maps a sampled chance outcome $c^{k+1}$, together with the afterstate $as^k$, to the next state.

The search in Stochastic AlphaZero constructs a tree that alternates between decision nodes and chance nodes, corresponding to states and afterstates, respectively.
During selection at a decision node, the algorithm applies a variant of the PUCT formula to choose an action. 
At a chance node, it samples an outcome according to the predicted chance probabilities. 

\textit{Stochastic Gumbel AlphaZero} (SGAZ)~\cite{kao_gumbel_2022} combines the algorithms of Gumbel AlphaZero and Stochastic AlphaZero to improve training efficiency in stochastic environments.
The effectiveness of the integrated method was demonstrated in \cite{kao_gumbel_2022}, where Stochastic Gumbel MuZero, an extension of SGAZ, was successfully applied to the popular puzzle game 2048.


\subsection{Evaluating Game Difficulty}
Recent studies leverage AI both to regulate challenges and to assess game variants.
One line of work is dynamic difficulty adjustment (DDA), where the agent's strength is adapted online to track the player's skill.
For example, AlphaDDA uses the standard AlphaZero architecture but adjusts the agent's strength based on the state value in real-time, for example, by changing the number of simulations or increasing dropout to make the agent weaker when needed.
Similarly, ROSAS and POSAS~\cite{demediuk_monte_2017} both modify the MCTS action-selection policy to match the opponent’s skill level.
In addition, a related study~\cite{tomašev_assessing_2020} to our work uses AlphaZero to analyze chess rule variants and quantify balance and decisiveness through expected scores and draw rates, as well as measures of opening diversity and approximate piece values.
This paper follows the same \textit{assessment via a strong agent} paradigm in the Tetris Block Puzzle environment.

\section{Tetris Block Puzzle} 

\begin{figure}[h]
    \begin{subfigure}[b]{\columnwidth}
        \centering
        \includegraphics[width=0.8\columnwidth]{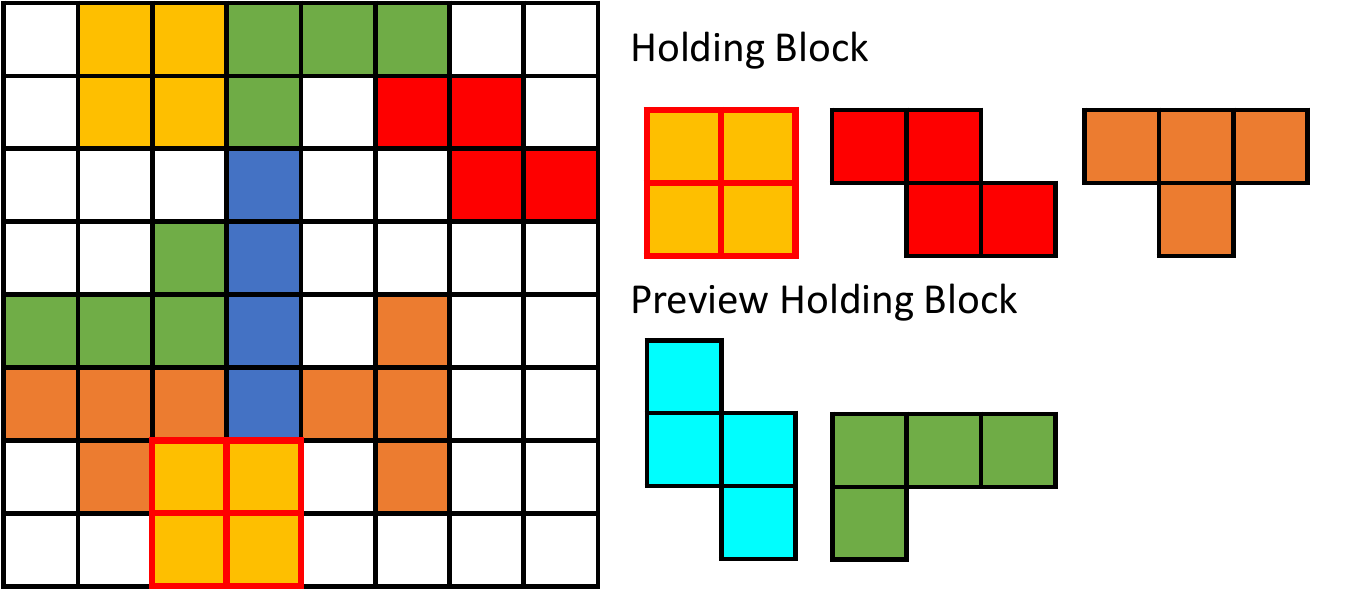}
        \caption{An example of the Tetris Block Puzzle game.}
        \label{fig:tbp-example}
    \end{subfigure}

    \vspace{0.5em}
    
    \begin{subfigure}[b]{\columnwidth}
        \centering
        \includegraphics[width=0.75\columnwidth, trim=0.5cm 0.5cm 0.5cm 0.5cm, clip]{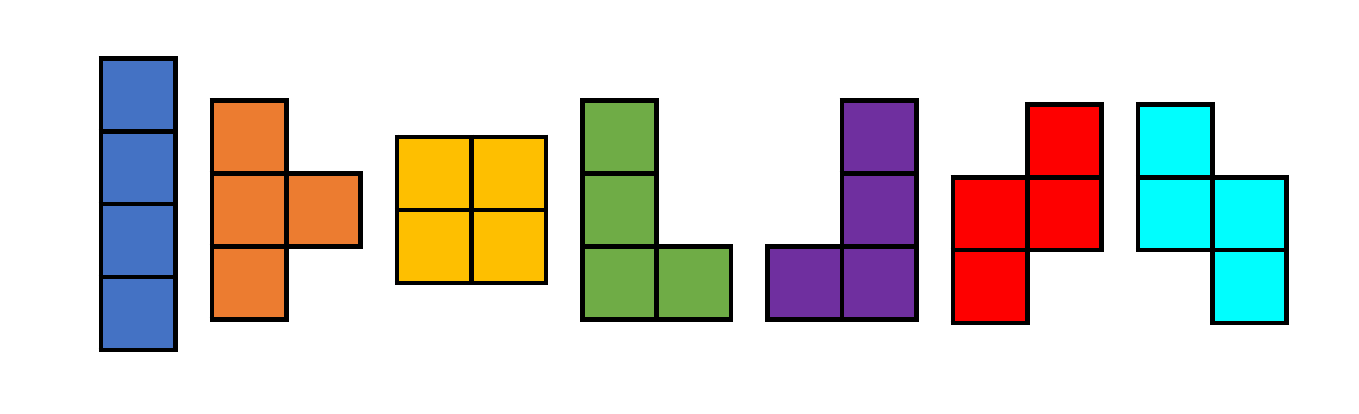}
        \caption{The standard tetromino blocks.}
        \label{fig:tbp-tetromino}
    \end{subfigure}

    \caption{
        The gameplay of Tetris Block Puzzle. (a) The player chooses an O-tetromino from three holding blocks and places it (highlighted with a red border), completing a vertical line.
        This clears the line and earns the player 1 point.
        After the O-tetromino is taken, the first preview holding block, an S-tetromino, becomes a new holding block for the next turn.
        Then, a new preview holding block is randomly selected from the standard blocks shown in (b) and may be rotated.
    }
    \label{fig:tbp-gameplay}
\end{figure}
Tetris Block Puzzle~\cite{website_tetrisblockpuzzle} is a single-player stochastic puzzle game where the player arranges blocks on an $8 \times 8$ grid to perfectly complete lines.
As illustrated in Figure \ref{fig:tbp-example}, tetromino blocks randomly appear during gameplay, and the player fits them into empty slots on the grid.
Once a line is fully placed, its blocks are cleared, and the player earns 1 point.
In order to analyze the difficulty of the Tetris Block Puzzle under different rule settings, we propose to modify the game rules in three specific ways as follows.

\noindent\textbf{Holding Block Rules.}
The \textit{holding blocks} refer to $h$ candidate blocks, from which the player chooses one to place in each turn.
When placing the block, it should be placed in the provided shape without any rotation.
After placing, a replacement block is automatically added to the set of holding blocks for the next turn.

\noindent\textbf{Preview Holding Block Rules.}
The \textit{preview holding blocks}, a feature not present in the classic game rules, are $p$ replacement blocks that will appear in the subsequent turns.
Specifically, the first preview holding block becomes a holding block after the player takes one step, and then a new preview holding block is randomly created at the end of the preview holding block sequence.

\noindent\textbf{Tetris Block Variants.}
In addition to the standard tetromino blocks shown in Figure \ref{fig:tbp-tetromino}, other types of Tetris blocks may be included to extend the challenge of the gameplay, e.g., pentomino blocks.
In this work, we examine this perspective by introducing the \textit{U-pentomino}, \textit{V-pentomino}, \textit{X-pentomino}, and \textit{T-pentomino} blocks, whose shapes are depicted in Figure \ref{fig:uvxt_blocks}.

Note that the rules of classic Tetris Block Puzzle are the configuration with $h=3, p=0$ and without any additional block.

\section{Experiments}
This subsection introduces the training process and then analyzes the modified game rules step by step.
To assess the difficulties of various Tetris Block Puzzle variants, we implement an extended game environment with adjustable rules using the MiniZero framework~\cite{wu_minizero_2025}.
In addition, to evaluate the difficulty of these game variants, we will use two metrics: \textit{Training Rewards} and \textit{Convergence Iterations}, as illustrated below.

\noindent\textbf{Training Rewards.} Represented by the average total rewards over the last 50 iterations before training is complete.

\noindent\textbf{Convergence Iterations.} Represented by the number of iterations required for the agent to consistently reach the maximum total reward over three consecutive iterations.

These two metrics allow us to evaluate the impact of various game rule changes on the difficulty and will be used for all of the remaining experiments. 
In addition, all the experiments were run on a machine with four 1080Ti GPUs.

\subsection{Training classic Tetris Block Puzzle}

\begin{figure}[ht]
    \centering
    \includegraphics[width=0.95\columnwidth]{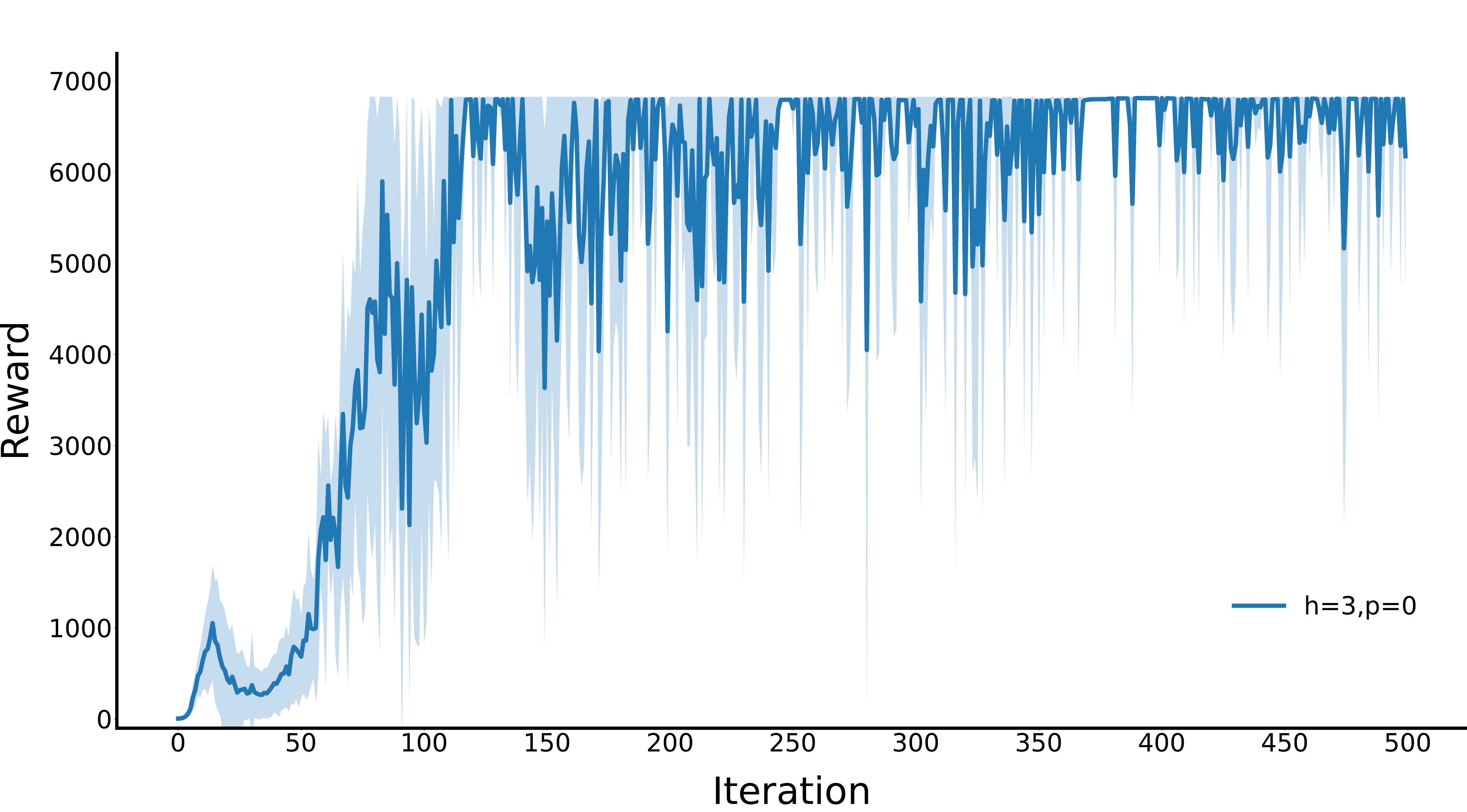}
    \caption{The training curve for the classic game with $h=3$ and $p=0$.}
    \label{fig:H3P0_training_curve}
\end{figure}

First, we train the model under the classic rules setting with $h=3$ and $p=0$, using SGAZ for 500 iterations to demonstrate that Tetris Block Puzzle can be effectively trained by SGAZ.
The average rewards per iteration are presented in Figure \ref{fig:H3P0_training_curve}.
The results show that SGAZ successfully learns the game, reaching an average total reward of 6544, which is very close to the maximum value set in the environment\footnote{To prevent a game from becoming endless, we set a maximum reward limit of 6750 points.}.


\subsection{Training Tetris Block Puzzle variants}
Next, following the same training setup, we provide game variants by modifying the number of holding blocks $h$, preview holding blocks $p$, and adding additional block types.

\subsubsection{Analyzing Holding Block Rules}

We evaluate game variants with varying only the number of holding blocks $h$, while keeping the number of preview holding blocks $p$ fixed at zero.

\begin{figure}[ht]
    \centering
    \includegraphics[width=\columnwidth]{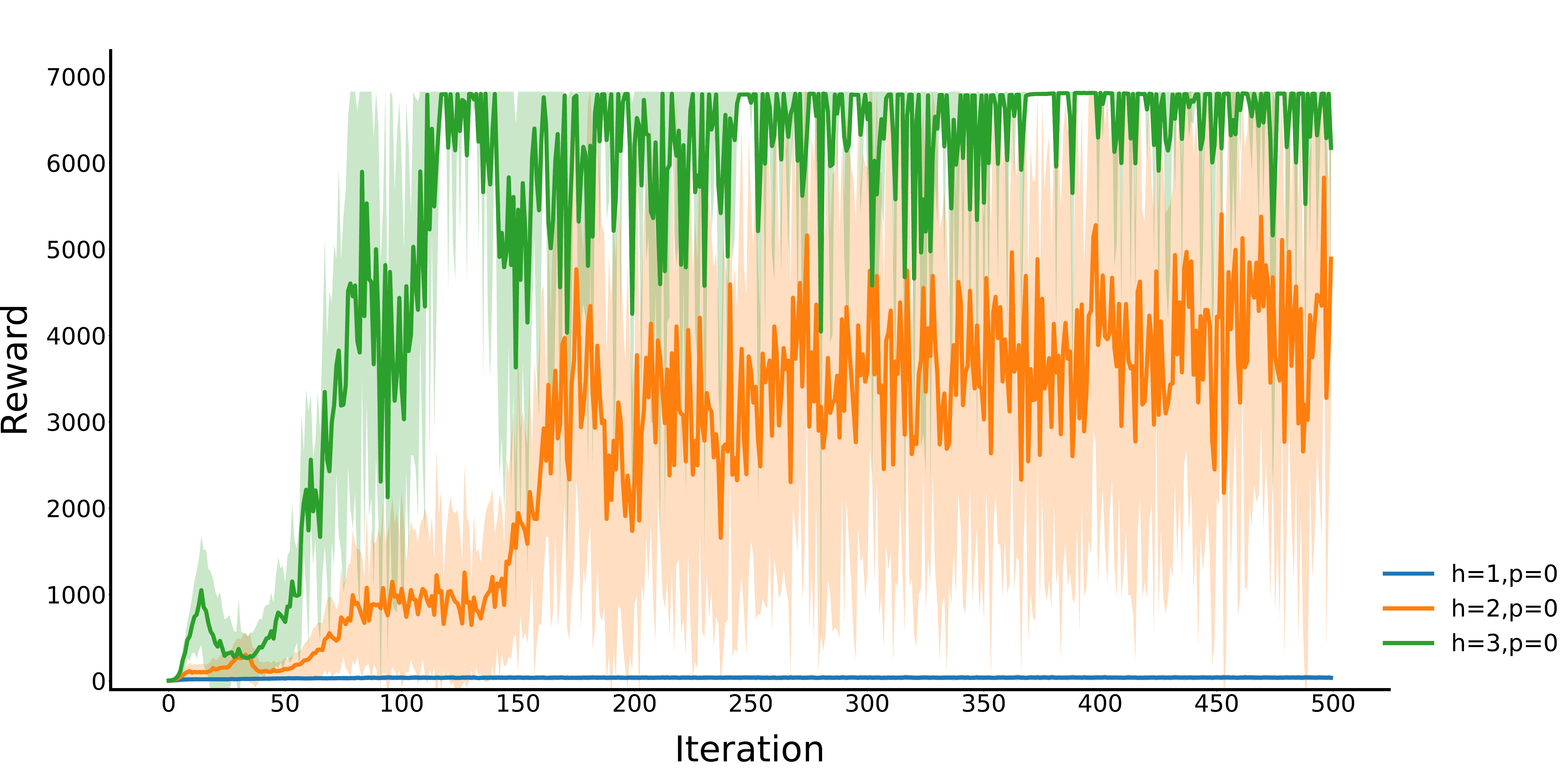}
    \caption{The training curve for $p=0$ with different \textit{h}.}
    \label{fig:H123P0_training_curve}
\end{figure}

\noindent\textbf{Training Rewards.}
The results in Figure \ref{fig:H123P0_training_curve} and Table \ref{tab:holding-blocks} show the reward during training. 
For the setting of $h=1$, the agent performs extremely poorly, indicating that this game rule is not suitable (too difficult), as SGAZ cannot play effectively. 
However, for other settings, the agent performs well, and even for the setting of $h=3$, SGAZ can achieve the maximum total reward.

\begin{table}[ht]
\caption{The training rewards of the different numbers of holding blocks.}
    \centering
    \setlength{\tabcolsep}{5pt}
    \begin{tabular}{lrrrr}
    \toprule
    $h$         &&  \multicolumn{1}{c}{1}  &   \multicolumn{1}{c}{2} &  \multicolumn{1}{c}{3} \\
    \midrule
    Reward     && 39.0 & 4126.1 & 6544.0 \\
    \bottomrule
    \end{tabular}
    \label{tab:holding-blocks}
\end{table}

\noindent\textbf{Convergence Iterations.}
Table~\ref{tab:convergence-speed-holding-blocks} reports the convergence iterations.
The results show that convergence iterations decrease as the number of holding blocks $h$ increases.
This means that with a larger $h$, the agent converges more quickly, and obviously, the game becomes easier.

\begin{table}[ht]
\caption{Convergence iterations for different numbers of holding blocks. The ``-'' mark indicates that the agent did not converge within training iterations.}
    \centering
    \setlength{\tabcolsep}{5pt}
    \begin{tabular}{lrrrr}
    \toprule
    $h$         &&  \multicolumn{1}{c}{1}  &   \multicolumn{1}{c}{2} &  \multicolumn{1}{c}{3} \\
    \midrule
    Convergence speed     && - & 160 & 61 \\
    \bottomrule
    \end{tabular}
    \label{tab:convergence-speed-holding-blocks}
\end{table}



\subsubsection{Analyzing Preview Holding Block Rules} 

Furthermore, we analyze the variants of fixing the number of holding blocks $h$ and varying only the number of preview holding blocks $p$ to assess their impact on the game.

\begin{figure}[h]
    \centering
    \includegraphics[width=\columnwidth]{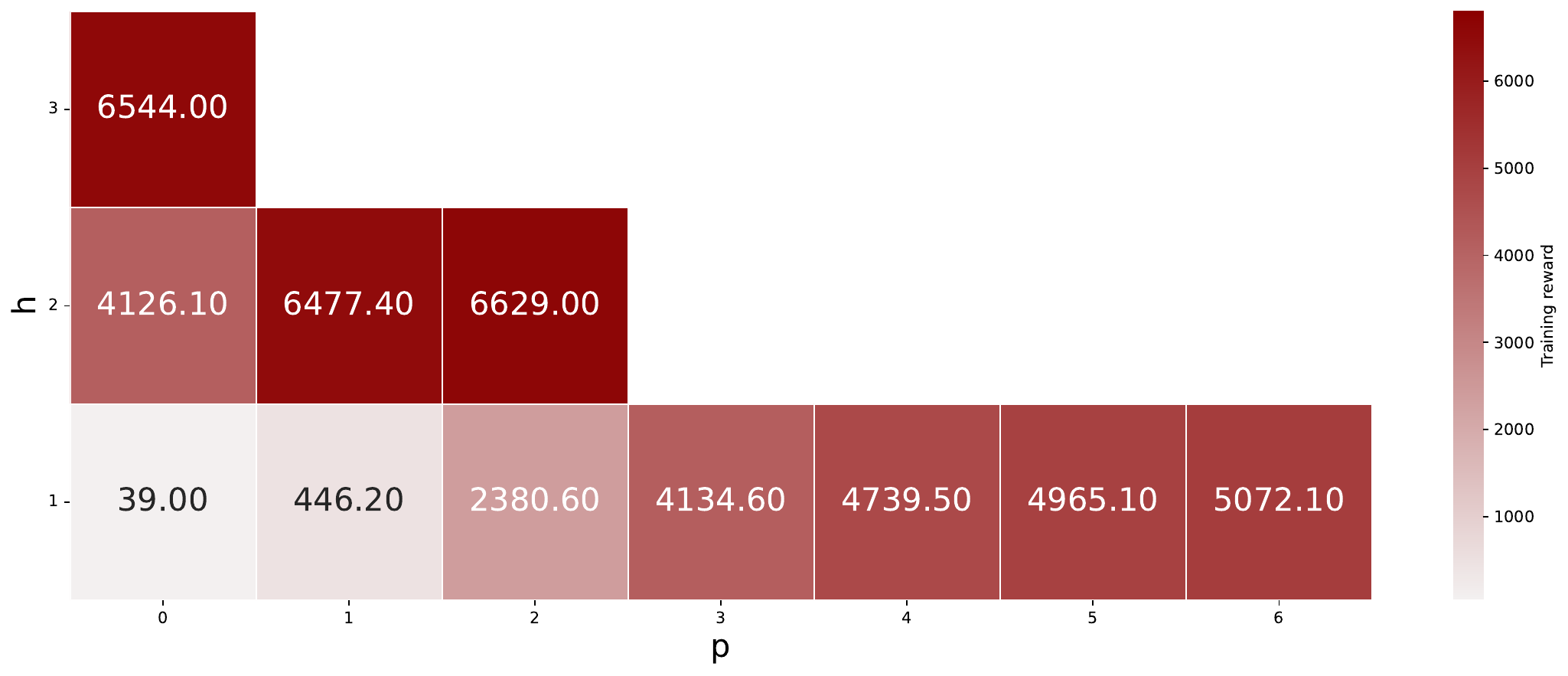}
    \caption{The training rewards impact of different numbers of preview holding blocks. The blank areas are expected to be optimal, as their preceding settings are already optimal.}
    \label{fig:hp_all_heatmap}
\end{figure}

\begin{figure}[h]
    \centering
    \includegraphics[width=\columnwidth]{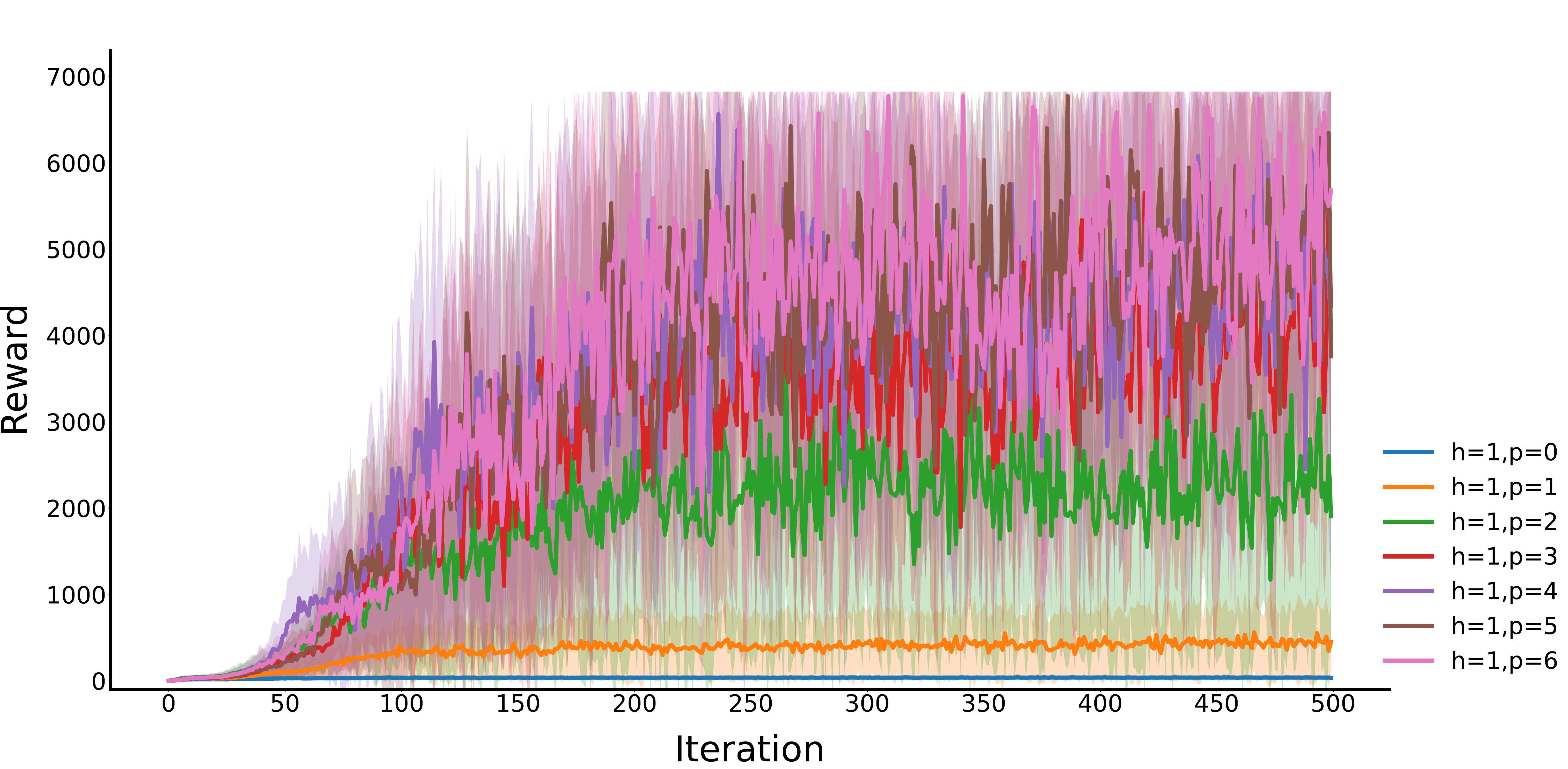}
    \caption{The training curve of different numbers of preview holding blocks under a fixed setting of holding blocks $h=1$.}
    \label{fig:H1Pall_training_curve}
\end{figure}

\begin{figure}[h]
    \centering
    \includegraphics[width=\columnwidth]{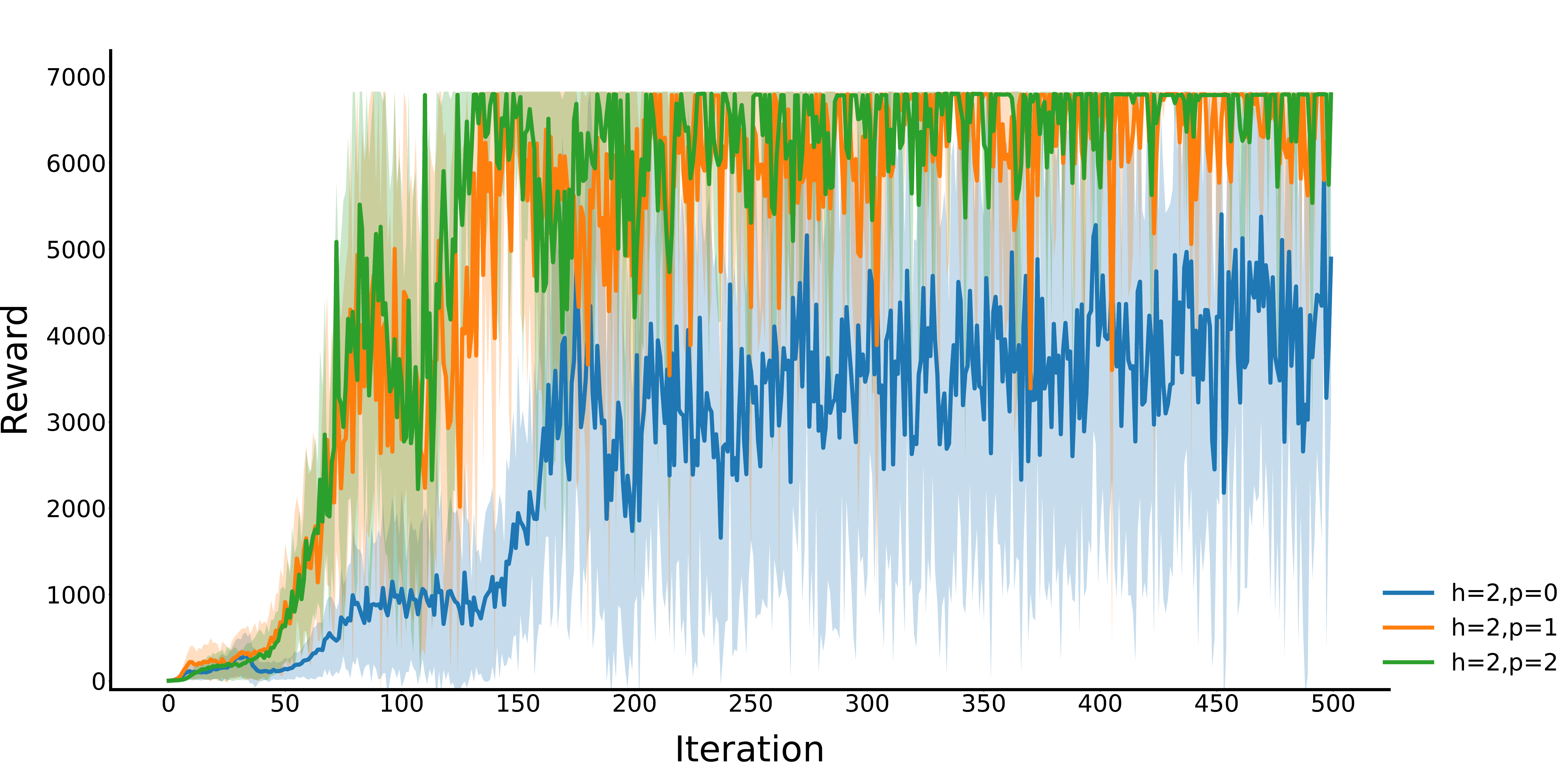}
    \caption{The training curve of different numbers of preview holding blocks under a fixed setting of holding blocks $h=2$.}
    \label{fig:H2Pall_training_curve}
\end{figure}

\noindent\textbf{Training Rewards.}
The results, summarized in Figure~\ref{fig:hp_all_heatmap}, \ref{fig:H1Pall_training_curve}, and \ref{fig:H2Pall_training_curve}, show that the training rewards increase as the number of preview holding blocks $p$ grows.
In addition, the results indicate that changing the number of preview holding blocks $p$ is less significant compared to changing the number of holding blocks $h$.
Specifically, with $h=1$ the training rewards increase slowly and converge at about 5000, while for other values of $h$ it can easily exceed this score.

\begin{figure}[ht]
    \centering
    \includegraphics[width=1\columnwidth]{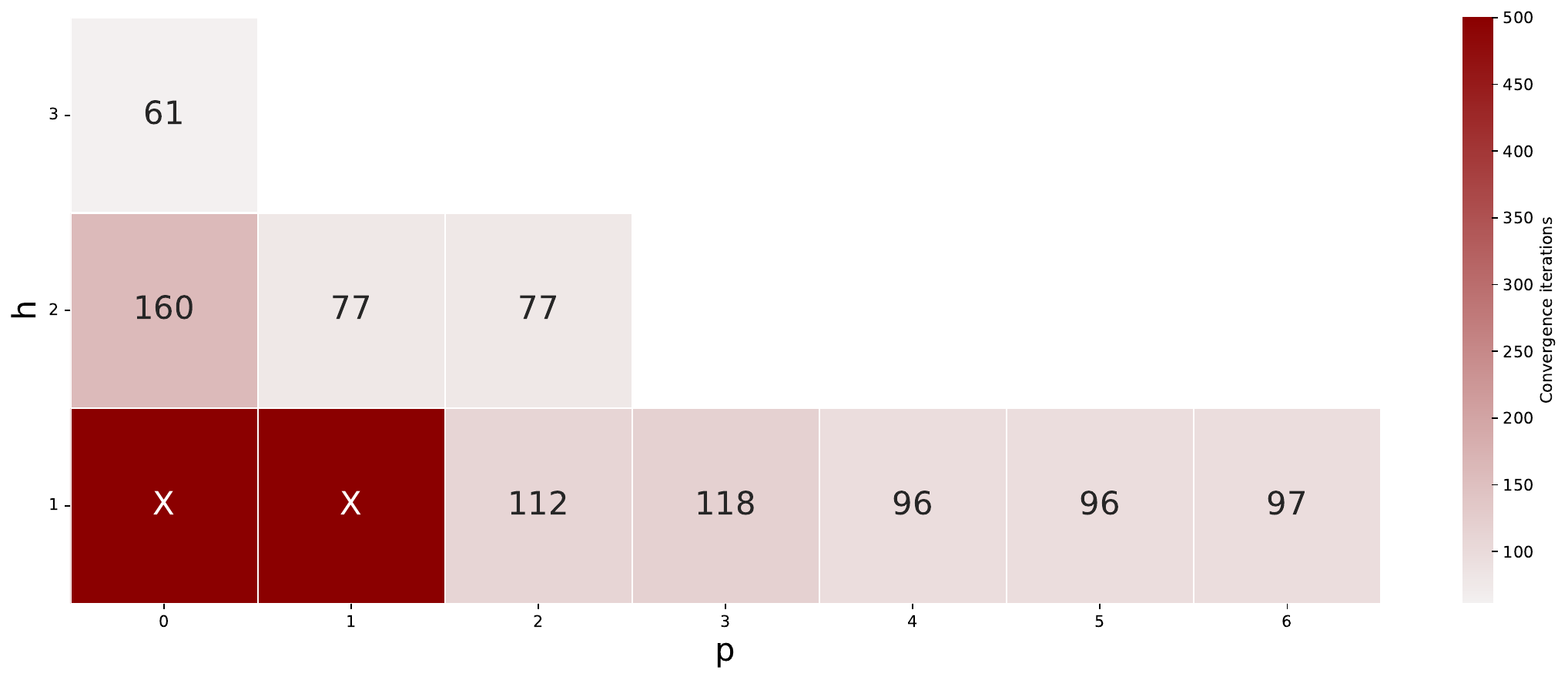}
    \caption{Impact of different numbers of preview holding blocks on convergence iterations. The mark of ``X'' area indicates that the agent did not converge within training iterations.}
    \label{fig:hp_all_convergence_heatmap}
\end{figure}

\noindent\textbf{Convergence Iterations.}
Figure~\ref{fig:hp_all_convergence_heatmap} reports the convergence iterations.
The results show the same situation as Figure~\ref{fig:hp_all_heatmap}, revealing that increasing $p$ slightly decreases the game difficulties.

\subsubsection{Analyzing Tetris Block Variants} 

\begin{figure}[ht]
    \centering
    \includegraphics[width=0.8\columnwidth]{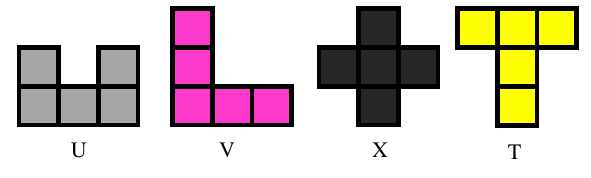}
    \caption{The U,V,X,T-pentomino blocks.}
    \label{fig:uvxt_blocks}
\end{figure}
\begin{figure}[ht]
    \centering
    \begin{subfigure}{0.8\columnwidth}
        \centering
        \includegraphics[width=\linewidth]{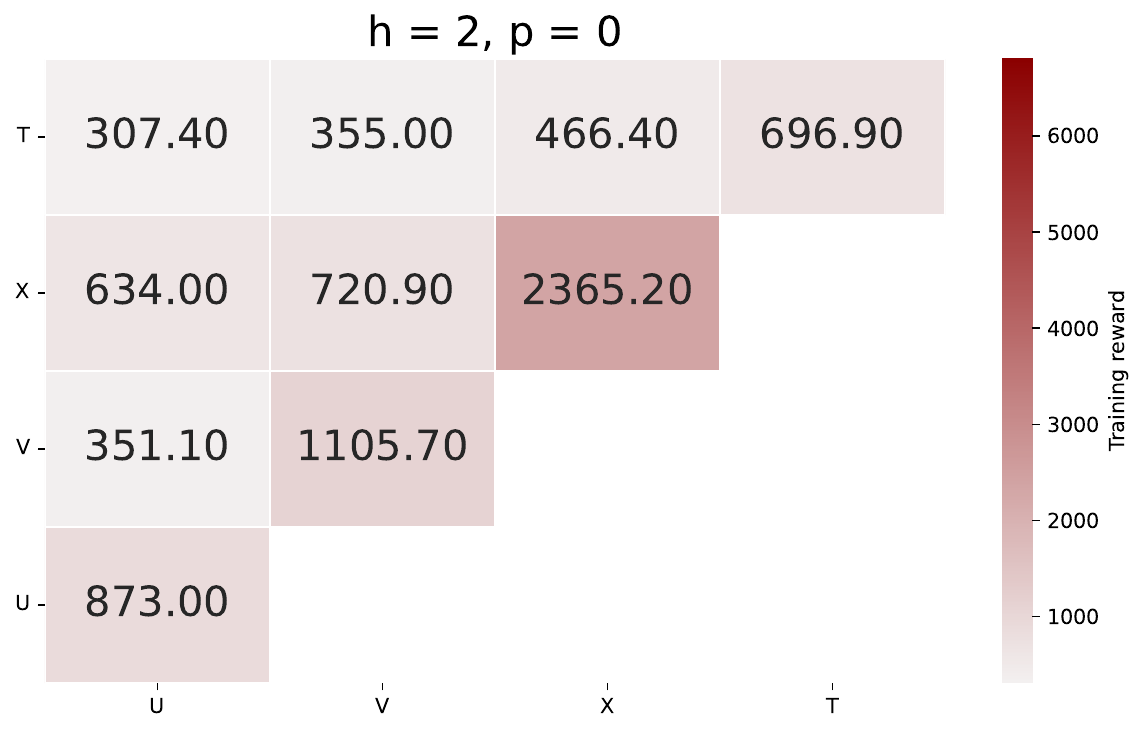} 
        \label{fig:pent:U}
    \end{subfigure}
  
    \vspace{0.5em}
  
    \begin{subfigure}{0.8\columnwidth}
        \centering
        \includegraphics[width=\linewidth]{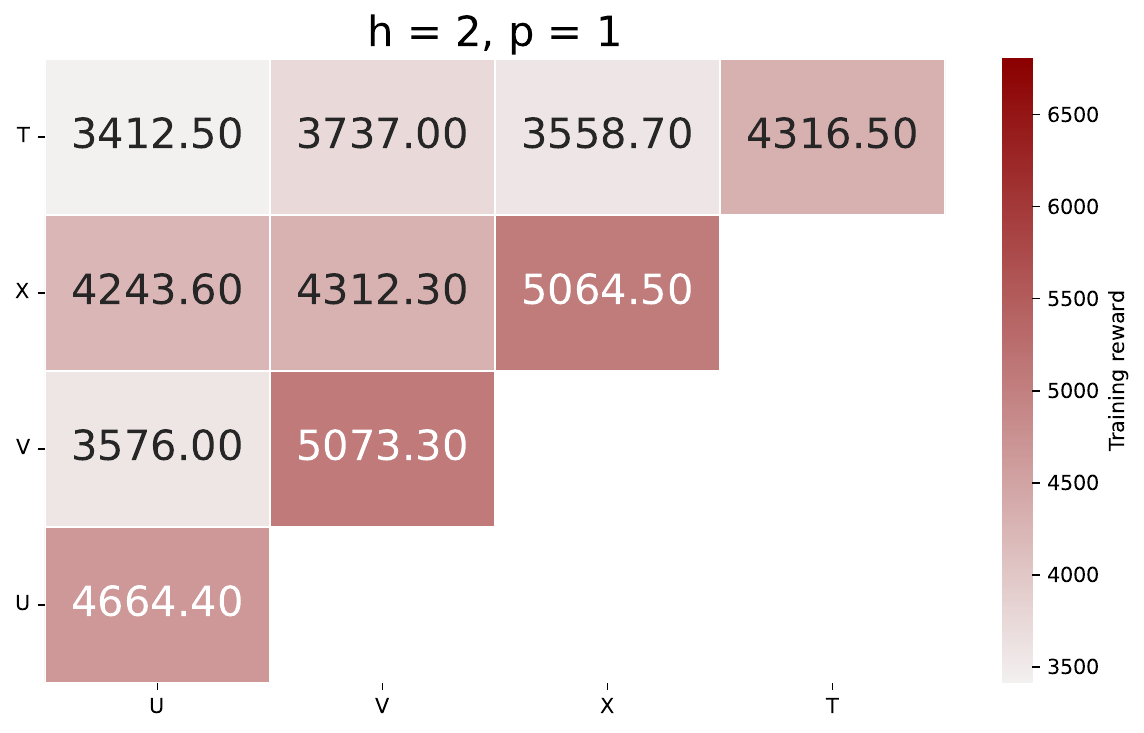} 
        \label{fig:pent:V}
    \end{subfigure}

    \vspace{0.5em}
  
    \begin{subfigure}{0.8\columnwidth}
        \centering
        \includegraphics[width=\linewidth]{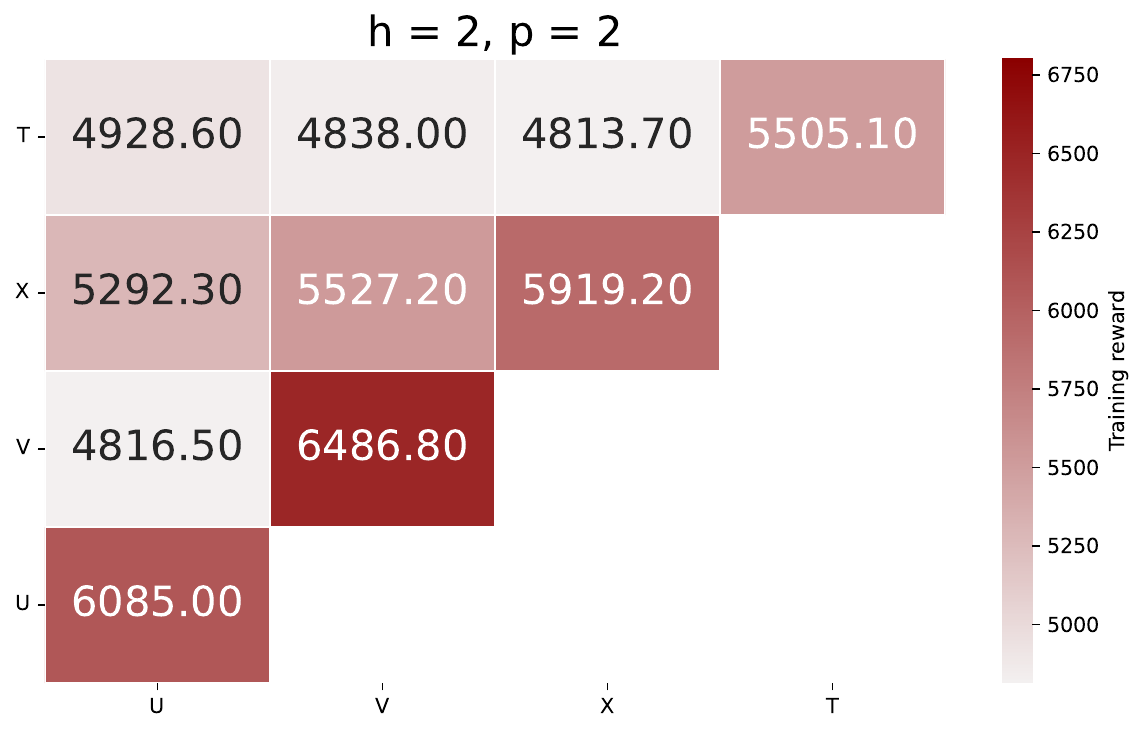} 
        \label{fig:pent:V}
    \end{subfigure}

    \caption{The training rewards impact of adding additional blocks.}
    \label{fig:UVXT_training_reward}
\end{figure}

Finally, we explore adjusting the difficulty by introducing additional blocks.
Using the U-pentomino, V-pentomino, X-pentomino, and T-pentomino blocks illustrated in Figure~\ref{fig:uvxt_blocks}, we assess the training performance of adding them under moderate settings of $h=2$ and $p\in \{0,1,2\}$.
For the additional blocks setting, we run experiments by adding either one or two blocks during training.
As shown in Figure~\ref{fig:UVXT_training_reward}, cells on the main diagonal (e.g., $(U,U)$ or $(V,V)$) correspond to runs with a single added block, while off-diagonal cells (e.g., $(U,V)$ or $(U,X)$) correspond to runs with two added blocks.

\noindent\textbf{Training Rewards.}
The results in Figure~\ref{fig:UVXT_training_reward} show that the training rewards decrease when any additional block is added, compared with Figure~\ref{fig:hp_all_heatmap}.
In addition, we can see that adding the T-pentomino has the largest impact, as the training rewards decrease more with the T-pentomino than with the other blocks.
This indicates that adding additional blocks would increase the game's difficulty.

\begin{figure}[ht]
    \centering
    \begin{subfigure}{0.8\columnwidth}
        \centering
        \includegraphics[width=\linewidth]{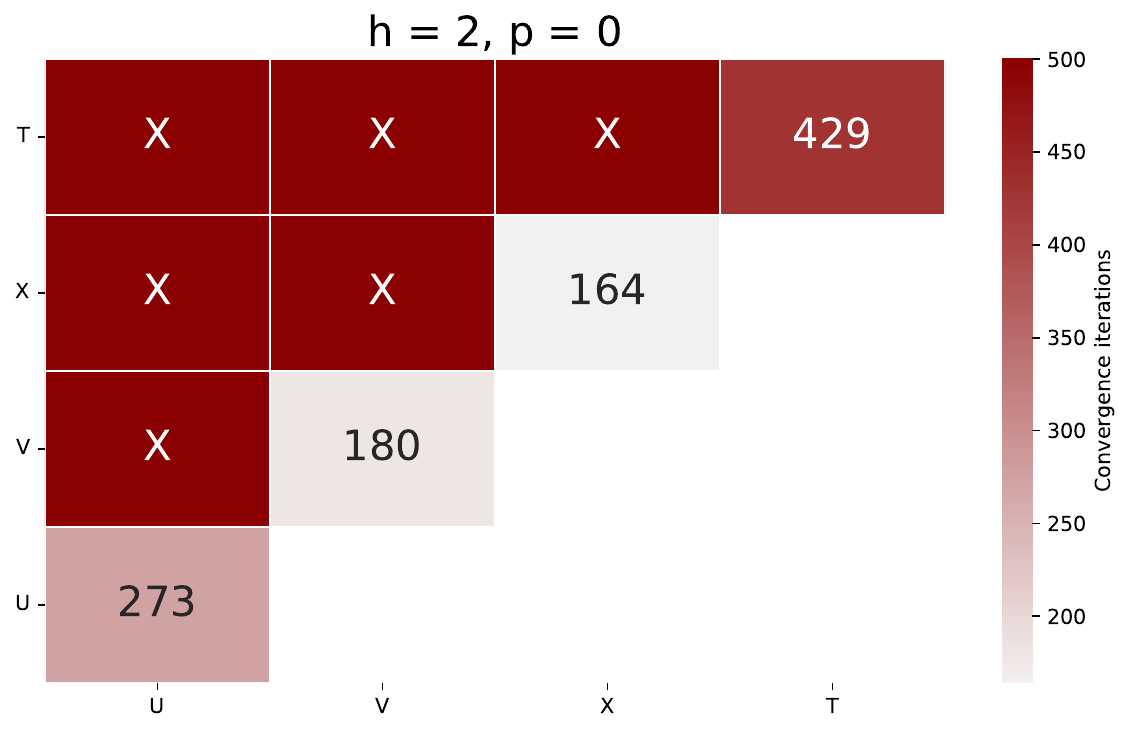} 
        \label{fig:pent:U}
    \end{subfigure}
  
    \vspace{0.5em}
  
    \begin{subfigure}{0.8\columnwidth}
        \centering
        \includegraphics[width=\linewidth]{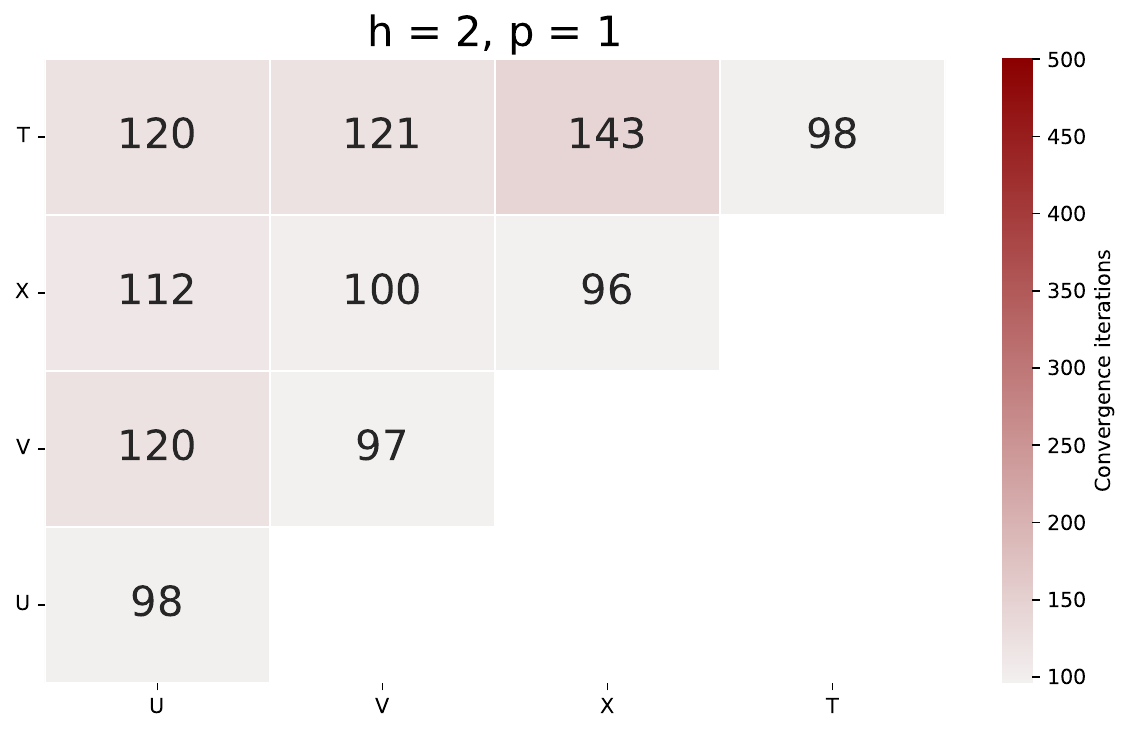} 
        \label{fig:pent:V}
    \end{subfigure}

    \vspace{0.5em}
  
    \begin{subfigure}{0.8\columnwidth}
        \centering
        \includegraphics[width=\linewidth]{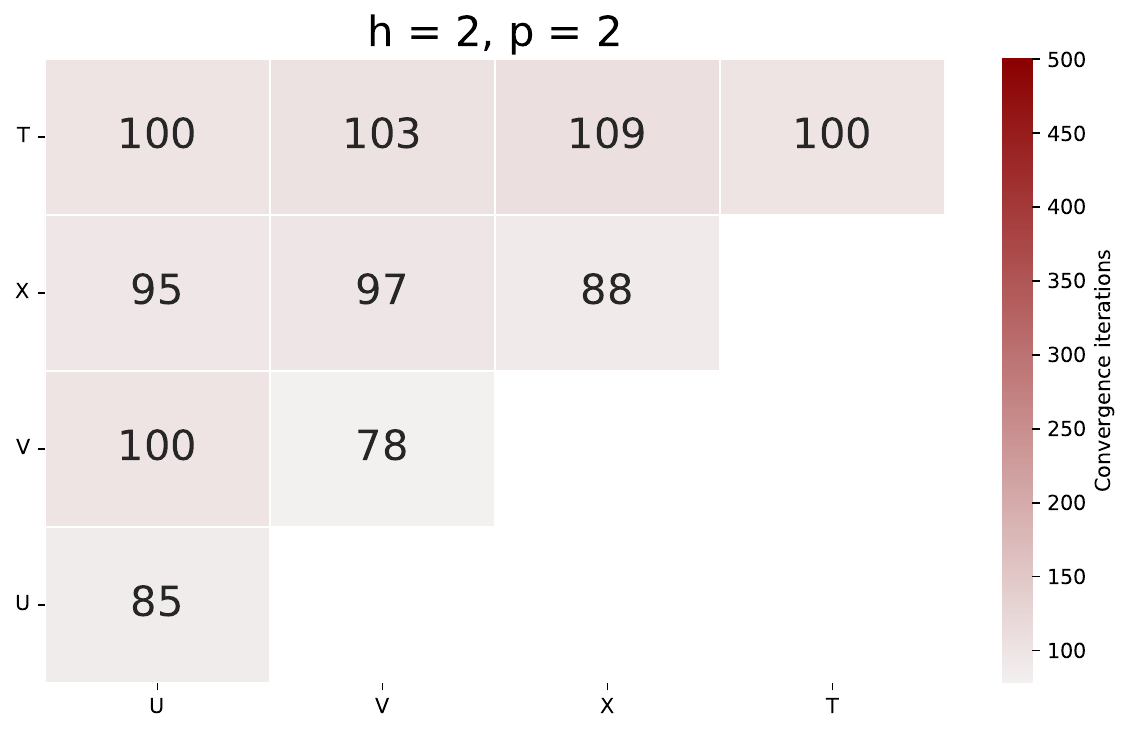} 
        \label{fig:pent:V}
    \end{subfigure}

    \caption{The convergence iterations impact of adding additional blocks.}
    \label{fig:UVXT_convergence_speed}
\end{figure}

\noindent\textbf{Convergence Iterations.}
Figure~\ref{fig:UVXT_convergence_speed} reports the convergence iterations.
Based on the result, adding an additional block will increase the game's difficulty.
We can observe that for the experiment setting with $h=2$ and $p=0$, all experiments involving the addition of two blocks did not converge during training.
This is a clear difference compared to the results shown in Figure~\ref{fig:hp_all_convergence_heatmap}, where convergence was easily reached.
In addition, the result still shows that the T-pentomino has the largest impact on the agent, as in every experiment where a single block is added, the T-pentomino always causes the slowest convergence speed.

\section{Discussion} 

In this paper, we assess the game difficulty of various Tetris Block Puzzle variants.
First, we formulate game variants based on three key settings: the holding block rules, the preview holding block rules, and the Tetris block variants.
We introduce two metrics to evaluate the difficulty of different game rules: Training rewards and Convergence iterations.
Additionally, we demonstrate that Stochastic Gumbel AlphaZero efficiently achieves near-optimal performance under mild settings, making it ideal for difficulty assessment.
Most importantly, our analysis indicates that increasing either the number of holding blocks or the number of preview holding blocks makes the game easier, though the former has a markedly stronger impact.
However, adding new block types significantly makes the game harder, especially for the T-pentomino block.
These insights refine our understanding of the game difficulty of Tetris Block Puzzle and provide a concrete reference for further research on stochastic puzzle games.

Looking ahead, we plan to explore more game variants, such as banning specific blocks from choosing, changing puzzle sizes, and introducing entirely new block types.
We also intend to conduct studies to evaluate how these variations affect human players' perceived challenge and enjoyment in the future.

\bibliography{gpw_2025.bib}
\bibliographystyle{style/ipsjsort-e}

\end{document}